\documentclass{article}
\usepackage{arxiv}
\usepackage[utf8]{inputenc}
\usepackage{rotating,multirow}
\usepackage{booktabs}
\usepackage{caption}
\usepackage{hyperref}
\usepackage[citestyle=authoryear,
]{biblatex}

\DeclareUnicodeCharacter{202F}{}

\newcommand{\ra}[1]{\renewcommand{\arraystretch}{#1}}

\author{ \hspace{1mm}Thomas Colligan \\
	Department of Computer Science\\
	University of Montana\\
	Missoula, MT 59801 \\
	\texttt{thomas.colligan@umontana.edu} \\
	\And
	\hspace{1mm}David Ketchum \\
	W.A. Franke College of Forestry and Conservation \\
	University of Montana\\
	Missoula, MT 59801\\
	\texttt{david2.ketchum@umconnect.umt.edu} \\
	\And
	\hspace{1mm}Douglas Brinkerhoff \\
	Department of Computer Science\\
	University of Montana\\
	Missoula, MT 59801\\
	\texttt{doug.brinkerhoff@mso.umt.edu} \\
	\And
	\hspace{1mm}Marco Maneta\\
	Department of Geosciences \\
	University of Montana\\
	Missoula, MT 59801\\
	\texttt{marco.maneta@mso.umt.edu} \\
}

\title{A Deep Learning Approach to Mapping Irrigation: IrrMapper-U-Net}
\date{March 4, 2021}

\hypersetup{
pdftitle={A Deep Learning Approach to Mapping Irrigation: IrrMapper-U-Net},
pdfauthor={Thomas Colligan, David G. Ketchum, Douglas Brinkerhoff, Marco Maneta},
pdfkeywords={irrigation mapping, computer vision},
}

\begin{document}

\maketitle
\section{Abstract}

Accurate maps of irrigation are essential for understanding and managing water resources. We present a new method of mapping irrigation and demonstrate its accuracy for the state of Montana from years 2000-2019. The method is based off of an ensemble of convolutional neural networks that use reflectance information from Landsat imagery to classify irrigated pixels, that we call IrrMapper-U-Net. The methodology does not rely on extensive feature engineering and does not condition the classification with land use information from existing geospatial datasets. The ensemble does not need exhaustive hyperparameter tuning and the analysis pipeline is lightweight enough to be implemented on a personal computer. Furthermore, the proposed methodology provides an estimate of the uncertainty associated with classification. We evaluated our methodology and the resulting irrigation maps using a highly accurate novel spatially-explicit ground truth data set, using county-scale USDA surveys of irrigation extent, and using cadastral surveys. We found that that our method outperforms other methods of mapping irrigation in Montana in terms of overall accuracy and precision. We found that our method agrees better statewide with the USDA National Agricultural Statistics Survey estimates of irrigated area compared to other methods, and has far fewer errors of commission in rainfed agriculture areas. The method learns to mask clouds and ignore Landsat 7 scan-line failures without supervision, reducing the need for preprocessing data. This methodology has the potential to be applied across the entire United States and for the complete Landsat record. 

\section{Introduction}

Irrigated cropland is responsible for approximately 40\% of the world’s crop production, despite only comprising 20\% of all cropland by area (\cite{FAO2020}). Irrigation already makes up about 85-90\% of all consumptive freshwater use globally (\cite{Deines2019}), and agriculture accounts for around 80\% of consumptive use in the United States (USDA ERS, 2019), of which use for irrigation is a large percentage. The FAO forecasts that by 2030 the world will have increased irrigated land by 34\% (\cite{FAO2020}). Despite the positive impact irrigation has on food production, it's associated with a range of environmental impacts ranging from aquifer and surface water depletion to soil salinization. The reliance of agriculture on irrigation makes spatially explicit knowledge of when irrigation is occurring and how it's changing over time essential to understanding and managing water resources. Maps of irrigation can also help inform climate and earth systems models, for example providing estimates of water diverted for agriculture to groundwater models. Most large scale maps of irrigation have been produced at low resolutions and/or rely on remote sensing data combined with national or subnational statistics (\cite{Thenkabail2009}, \cite{Siebert2015}, \cite{Meier2018}, \cite{Salmon2015}, \cite{Brown2014}, \cite{Xie2019}, \cite{Ozdogan2008}). Agricultural statistics collected by governments or NGOs often do not capture high resolution temporal trends in irrigation due to labor-intensive collection of data and can contain bias, especially if there are political or other incentives involved in reporting irrigation. 

Methods based solely on census data are limited in resolution by the administrative boundaries used to collect the data. For example, the United States' National Agricultural Statistics Survey (NASS), which reports irrigated area by county in the US, is compiled every 5 years by self-reported survey. The statistics are generally regarded to be an accurate characterization of crop type and irrigation status for lands across the US, but could be biased due to incentives to over- or under-report (\cite{Ozdogan2008, Deines2019}). This occasional snapshot of irrigated area does not pick up short term fluctuations in irrigated agriculture. In addition, NASS excludes irrigated areas that produce less than \$1,000 (\cite{NASS}) in crops, slightly biasing its surveys. Definitions of irrigation may also depend on the agency collecting the data, and may provide area equipped for irrigation, not actual irrigated area (\cite{Ozdogan2008}). Irrigated area and area equipped for irrigation may be significantly different in any given year due to fluctuations in crop prices or rotation. The spatial resolution (ranging from 500m to 5 arcsec) of many existing maps of irrigation of the globe and the continental United States (CONUS) makes the identification of small or partially irrigated areas difficult. The low resolution of previous maps of irrigation on large scales is due to the extensive computing resources required to process satellite data and the effort required to gather ground truth samples. However, the increasing ability of researchers to work with large data sets has made mapping irrigation at high resolution possible.

Google Earth Engine (GEE) (\cite{Gorelick2017}) has aggregated remote sensing data in one place and provided free access to their compute infrastructure, helping researchers increase the resolution and scale of remote sensing mapping approaches. The introduction of GEE to the remote sensing community has made mapping irrigation at high resolution on large temporal and spatial scales possible without relying on local computing power. This shift in technology has led to multiple efforts at mapping irrigation using Landsat data on large scales and high resolutions (\cite{Deines2017, Deines2019, Xie2019, Ketchum2020}). These efforts all rely on Random Forest (RF) models and novel ground truth datasets to map irrigation across regions ranging from the High Plains Aquifer to the entire continental United States (CONUS). RFs are common in the remote sensing community for their flexibility, interpretability, ease of use, and their ability to handle complicated high-dimensional data. (\cite{Deines2017, Deines2019, Xie2019, Ketchum2020}) are the only high resolution (30m) irrigation maps that cover large regions in time and/or space in the United States. LANID (\cite{Xie2019}), is a Landsat-based irrigation map that uses a semi-automatic method of generating training data to train RF models that classify irrigation across CONUS at 30m. The semi-automatic generation of training data uses estimates of irrigated area by county from MIrAD-US (\cite{Brown2014}) (a 250m irrigation map that matches NDVI- derived estimates of irrigation area directly with NASS statistics by county) to threshold maps of satellite-derived spectral indices until the area reported by the maps and MIrAD match. The thresholded maps serve as areas in which (\cite{Xie2019}) sample points to train RF models, which are used to classify land as irrigated or non-irrigated. The LANID method is based on county level statistics through the reliance on training data generated by matching MIrAD-US. IrrMapper (\cite{Ketchum2020}) and AIM-HPA (\cite{Deines2019}), take a different approach and train their random forest models with hand-labeled training points, producing estimates of irrigated land independent of NASS. IrrMapper covers the western United States and AIM-HPA extends across the High Plains Aquifer. LANID and AIM-HPA both mask their classification product to an ancillary land cover map such as the Cropland Data Layer (CDL). IrrMapper (and our method) removes this postprocessing step, instead using the CDL as an input feature. The RF methods above (LANID, IrrMapper, and AIM-HPA) all report overall accuracy greater than 90\% and agree well with NASS statistics. 

Part of RFs widespread use is their free implementation in GEE. However, the state of the art in image segmentation (of which irrigation mapping is a subset) is achieved through the application of deep convolutional neural networks (CNNs). CNNs can match and even exceed human performance in various computer vision tasks. CNNs incorporate learned representations of images on many scales when making predictions, picking up patterns and contextual information. This is especially important in irrigation mapping, where the difference between a non-irrigated field and irrigated field may not be solely in spectral information but in the shape of the field or surrounding context. RF models, in contrast, have to rely on handcrafted features that capture spatial context, or otherwise discard spatial information in their classification. Another advantage of CNNs is that they can use dense pixel-wise labels while RF models have to pick and choose which points they use for training. CNNs have been used in many domains of remote sensing, from crop type mapping and land cover classification (\cite{Zhong2019, Rußwurm2018}), to change detection (\cite{Peng2019, Janalipour2017}). Despite their proven utility in remote sensing applications, there are only a few examples of CNNs directly applied to producing maps of irrigation, especially at large scales. The main reason for this deficit is the limited availability of training data (\cite{Deines2017}) and high computational costs. Training deep neural networks effectively requires thousands of training, testing, and validation samples as well as specialized computer hardware (e.g. graphical processing units, GPUs). Evaluating these models over a large geographic region at high resolution can also cost-prohibitive, as CNNs are often computationally expensive. Despite the disadvantages, the accuracy of CNNs routinely surpasses that of RF-based approaches because of their ability to learn explanatory features without supervision. The only use of deep learning to produce maps of irrigation to our knowledge at the time of writing is the work by (\cite{Zhang2018}), and (\cite{Saraiva2020}). Both methods use CNNs to map center pivot irrigation patterns over the state of Colorado and the Cerrado Biome of Brazil, respectively. These methods both relied on intensive hand-labeling of center pivot irrigation structures in satellite images. Both achieved overall accuracy of greater than 99$\%$, and limited their study area to a relatively small region in time and space (20000km$^2$, 1986 and 2000, and 37000 km$^2$, 2017, 2018 and 2019, respectively).

Our objective was to produce a neural network model and framework that is easy to use and achieves better results than current approaches to mapping irrigation on a large spatial and temporal scale. We use novel ground truth data (a subset of the data used in \cite{Ketchum2020}) to train an ensemble of neural networks to label Landsat pixels as irrigated or non-irrigated in the state of Montana for years 2000-2019. Unlike other products we produce pixel-wise uncertainties for our maps of irrigation. We conduct a spatially explicit pixel-wise comparison between methods (IrrMapper, U-Net ensemble, and LANID) using the novel ground-truth data. Our method outperforms other methods in overall accuracy and precision (also known as errors of commission), despite limited feature engineering and the omission of a crop mask like the CDL. Finally, we compare our method to NASS statistics and Final Land Unit (FLU, \cite{FLU2020}) classification data from Montana. We find better agreement for most years between total irrigated area in Montana calculated with our approach and NASS data compared to existing maps of irrigation. We also report on the qualitative performance of U-Net, LANID, and IrrMapper, describing the successes and failures of each product.

\section{Methods}

\subsection{Study area}

\begin{figure}[h]
    \centering
    \includegraphics[width=\textwidth]{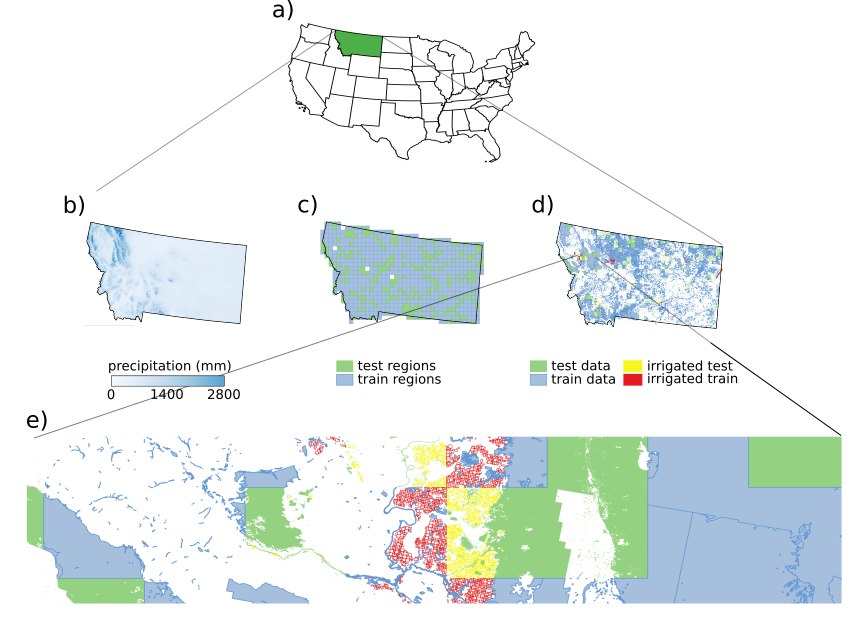}
    \caption{Study area and data used in this analysis. Subfigure b) indicates the mean annual precipitation for the years 2000-2019, c) shows the test and train regions spread over the state of Montana, d) the training and test polygons. A zoomed in view of the training and test polygons is shown in subfigure e. Irrigated training polygons are in red and test polygons are in yellow. For ease of visualization, the unirrigated and uncultivated class are aggregated into one class.}
    \label{fig:data_description}
\end{figure}

We focus on the state of Montana (MT), United States, which has an area of $\sim$ 380,000$km^2$. West of the Continental Divide, Montana is generally wetter and more temperate than area east of the CD, which is drier and windier (\cite{WaterPlan2015}). Montana's mean annual precipitation for 2000-2019 is shown in subfigure b) of Fig. \ref{fig:data_description}. Irrigation in the state of Montana is mostly applied by diverting surface water (99\%) and groundwater pumping (1\%) (Lonsdale et al., 2020), and comprises $\sim$68\% of consumptive water use (\cite{WaterPlan2015}) in the state. 

\subsection{Labels}

We used the same data in (\cite{Ketchum2020}), restricted to the state of Montana. These data are vector labels partitioned into three classes: irrigated, unirrigated, and uncultivated. All field labels were created through manual interpretation of satellite images. Labels were either hand-labled for this study or collected from online sources. Each vector feature in the irrigated label set was vetted for accuracy using high-resolution imagery and ancillary data like precipitation. Unirrigated labels cover predominately dryland (rainfed) agriculture, and include some fallow fields. The uncultivated labels contain land which is not used for crop cultivation, like rivers, lakes, wetlands, forest, and prairie. Irrigated and fallow field labels were used for 2003, 2008-2013, and 2015 to represent a range of water availability, using yearly precipitation as a proxy \footnote{Using NOAAs Climate At A Glance tool, annual precipitation was plotted for 2000-2019  (\url{https://www.ncdc.noaa.gov/cag/statewide/time-series/24/pcp/ann/1/2000-2021?base_prd=true&begbaseyear=1901&endbaseyear=2000}). Years were selected in this time range to build training data if they had diverse amounts of precipitation.}. Areas labeled as unirrigated or uncultivated were assumed to stay the same class over the years studied. Overall, training data was used to train the model for years 2003, 2008-2013, and 2015. 

To ensure no overlap between training and validation data, a grid of 776 23040 m\textsuperscript{2} non-overlapping squares was spread over Montana and then split into subsets of 582 and 194 (80, 20\% splits, subfigure c in Fig. \ref{fig:data_description}). The vector labels described above were then split into training and validation sets based on this grid.

\subsection{Features}

We used Landsat Tier 1 Surface Reflectance data from Landsat missions 5, 7 and 8, acquired through GEE. The Landsat mission is the longest continuous record of satellite images and minimal effort is required to standardize across missions. Landsat images of the earth are also captured frequently, with approximately 8 days between subsequent observations across all satellites. The bands used for analysis were blue, green, red, NIR, SWIR 1, and SWIR 2. We did not perform any cross-calibration between Landsat sensors, nor did we mask clouds or cloud shadows. Cloud masks were not used in order to force the machine learning algorithm to learn a robust representation of irrigation that ignores clouds.

To standardize data across years, we took temporal means of images captured over a 32-day period, assuming that the temporal mean value of a Landsat band stays relatively stationary over time. All available Landsat images starting on May 1st and extending for 192 days were collected and sorted by time using GEE. For each non-overlapping 32-day segment within the 192 days, the temporal mean of all images collected during that segment was computed, and the resulting mean image was stored and used as a feature to the model. The final features that were ingested into the model were six temporal mean images (each comprised of 6 bands: blue, green, red, NIR, SWIR 1, and SWIR 2) sorted by time.

We treat each band as a separate feature without temporal information, which is standard for image segmentation models. The 32-day time period was chosen because it represented a trade-off between temporal resolution and number of input features. Taking the mean over a longer time period could suppress important information for determining irrigation. A shorter aggregation period would cause the number of input features to rise, increasing computational and storage needs. The training and validation data are not processed in any way to compensate for missing data due to the Landsat-7 scan-line failure. The failure occurred on May 31, 2003, so the scan-line failure was present in all of the training and validation data. Missing data due to the failure was noticeable in the data upon inspection. The noisy data serves as data augmentation, as the machine learning algorithm has to learn a representation of irrigation that ignores scan line artifacts.

\subsection{Model description}

\begin{figure}[h]
    \centering
    \includegraphics[width=0.9\textwidth]{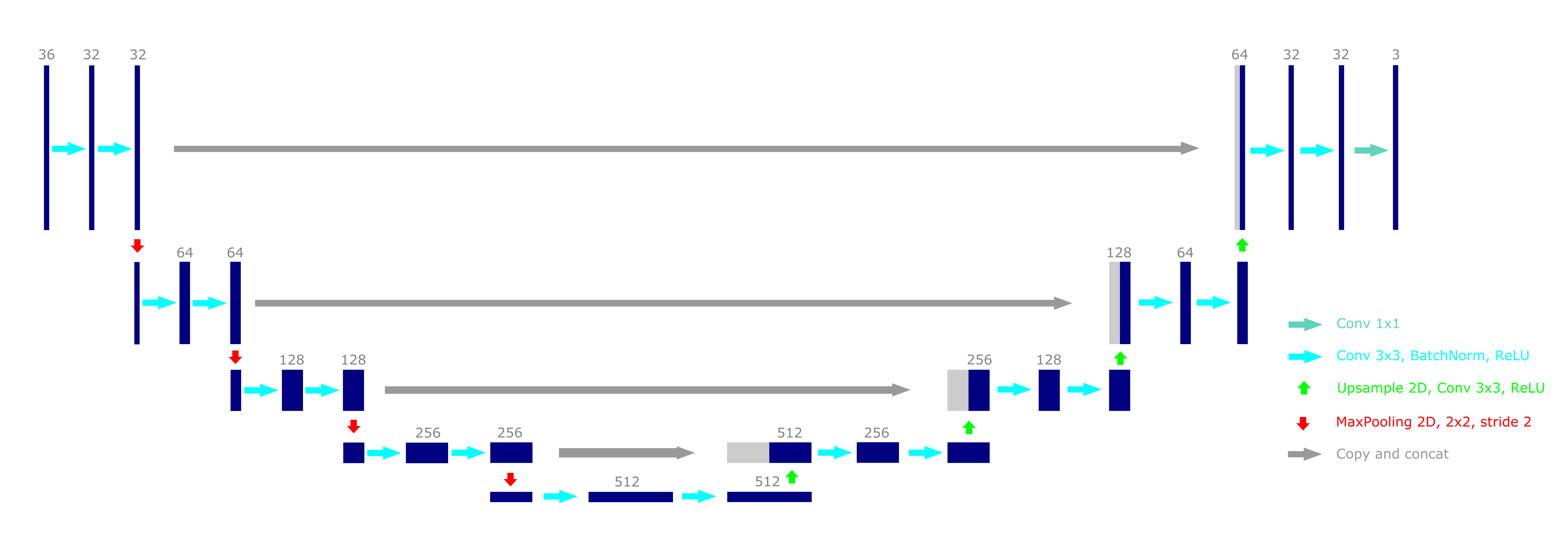}
    \caption{Neural network architecture used in this study. Blue bars indicate convolutional filter blocks. Numbers above blocks represent the number of filters in the block. Arrows represent operations on the blocks.}
    \label{fig:model}
\end{figure}

The convolutional neural network used to predict irrigation was a variant of the U-Net model (\cite{Ronneberger2015}). The U-Net architecture was chosen because it is a tried-and-true method for image segmentation and because it is a fully-convolutional neural network (FCNN). FCNNs are noted for their fast inference time compared to pixel-wise methods (\cite{Shelhamer2017}), which means maps of irrigation can be produced in comparatively little time on a personal computer. See the Appendix for more details on convolutional neural networks. A diagram of the model is shown in Fig. \ref{fig:model}, and coincidentally is very similar to the architecture used in (\cite{Saraiva2020}) to identify center pivot irrigation. 

The model was composed of two main parts: A contracting path that downsamples the image to extract low-level features and spatial context, and an expanding path that incorporates low-level information from the contracting path to create high resolution predictions. We followed each convolutional layer with a batch normalization layer (\cite{Ioffe2015}) and used the ReLU activation function. We used weight decay with a coefficient of 0.001 on all convolutional layers. Unlike the original U-Net implementation, we used 2-d upsampling instead of transposed convolution to increase resolution. Each convolution is zero-padded. The final layer in the model is a convolutional layer with three filters of kernel size 1 followed by a softmax activation function. The softmax layer produces three predictions of class probability per pixel that sum to 1. In the original implementation of U-Net, the number of filters at each convolutional layer starts at 64 and doubles after each max-pooling layer. We use the same approach, but the first convolution in our U-Net implementation consists of only 32 filters, reducing the overall number of parameters from 32 million to 8 million. Fewer parameters means the model took less time to train and evaluate.

Many applications of U-Net perform segmentation on single images, with either 3- or 4-channels. Using a single image for land cover classification would discard important temporal information, which is especially important for classifying irrigation (\cite{Ozdogan2008}). The idea of concatenating features from different times and feeding them into a model is not new (\cite{Ozdogan2010}) in irrigation mapping, yet has not been applied in a fully-convolutional neural network. Accordingly, we concatenate temporal mean images taken over a growing season into a 36 channel image. The 36 channel image consists of 6 images with 6 channels collected at different times.

\subsection{Training details}

We trained 10 models on random subsets of the training data as model ensembles are shown to do better than their constituents and can provide a measure of predictive uncertainty (\cite{Breiman1996}). Each model was trained for 100,000 gradient descent steps (batch size 25) with the Adam (\cite{Diederik2014}) optimizer with default $\alpha$, $\beta_{1,2}$ parameters. We used random majority undersampling to roughly balance examples from each class per batch. The initial learning rate was 0.001, and was step decayed every 25,000 steps by 60\%. Each model took around 24 hours to train on a single NVIDIA GTX 1070 GPU. No hyperparameter tuning was performed since there was no test set. 

\subsection{Analysis Pipeline}

Maps of irrigation were produced for the years 2000-2019 over Montana. We used the overlap-tile strategy in (\cite{Ronneberger2015}) to produce seamless predictions over the region of interest for each model. To reduce data storage requirements, the output of the models was scaled to the range 0-255 (from 0-1) and converted to an unsigned 8-bit integer. The median predicted probability for each pixel in each band was calculated and the largest median value in each class was used as the final classification. To quantify ensemble uncertainty, we created maps of inter-quartile range (IQR) of the predictions for each pixel. A smaller IQR meant that models agreed well on the classification of a given pixel, and a large IQR indicated high disagreement.

The vector labels used to train the model were collected at higher resolution (centimeter to meter resolution imagery) than the Landsat resolution, resulting in separate fields that were close to one another becoming one shape when rasterized. The models learned to label small roads between fields as irrigated.

\begin{figure}[h]
    \centering
    \includegraphics[width=\textwidth,height=\textheight,keepaspectratio]{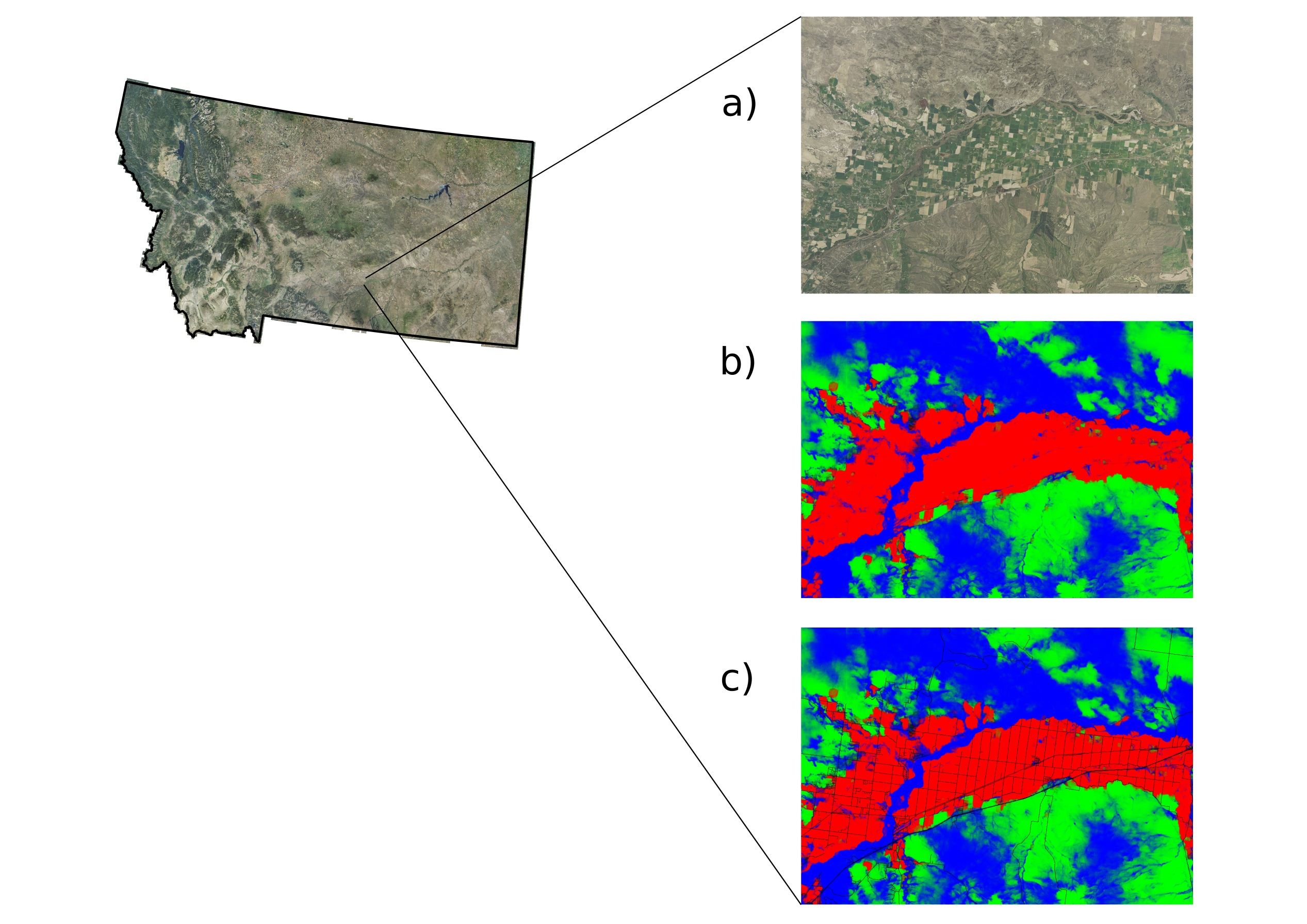}
    \caption{a) NAIP imagery of irrigation near Billings, Montana (2013). b) corresponding unmasked model predictions. Red indicates irrigation. c) Irrigation predictions masked to roads shapefile.}
    \label{fig:masked_to_roads}
\end{figure}

We used human-labeled data to post-process roads out of model predictions. We masked the pixels in the final evaluated rasters with features that describe all roads in Montana\footnote{downloaded from \url{https://geoinfo.msl.mt.gov/home/msdi/transportation/}}. An example of the initial predictions and the masked predictions is shown in Fig. \ref{fig:masked_to_roads}.

\subsection{Accuracy analysis}

We conducted a pixel-wise comparison of LANID, IrrMapper, and U-Net on the validation data. The validation data consisted of hundreds of millions of pixels, with $\sim$100$\times$ more non-irrigated pixels than irrigated pixels. F1-score was used to quantify errors of omission and commission on irrigated labels. F1-score aggregates the number of true positives, false positives, and false negatives into one number, defined as f1=$\frac{TP}{TP+\frac{1}{2}(FP+FN)}$. An alternative formulation of f1-score relies on precision and recall (also known as errors of commission and omission, respectively), f1=$\frac{2*P*R}{P+R}$, where P=$\frac{TP}{TP+FP}$, and R=$\frac{TP}{TP+FN}$. We use f1-score, precision, and recall to quantify the performance of the various methods (LANID, IrrMapper-RF, and the U-Net ensemble) on the irrigated class. 

To the best of our knowledge, IrrMapper, MIrAD-US, and LANID are the only other high spatial and temporal resolution irrigation maps that cover the state of Montana. MIrAD-US is only available for years 2002, 2007, 2012, and 2017, as it is directly parameterized using NASS statistics (and as such agrees very well with NASS predictions of irrigated area by county). We did not produce pixel-wise assessments of MIrAD-US maps since they are 250m resolution compared to the 30m of LANID, IrrMapper, and U-Net. IrrMapper covers the range 1986-2019, and we were able to access LANID for 2008-2013. To compare between models, we aggregated U-Net classes unirrigated and uncultivated into a single class (non-irrigated), as the other products only assign a binary indicator to each pixel, and evaluated each models performance on the validation split.

To assess the difference between model prediction of irrigated land and agricultural statistics data, we compared U-Net results to two different estimates of irrigated area, NASS and FLU. NASS statistics were available for the years 2002, 2007, 2012, and 2017, shown in Fig. \ref{fig:unet_vs_nass_by_county}. FLU is a vector dataset produced by the Montana Department of Revenue for agricultural property valuation that classifies private agricultural land into one of six uses: fallow, hay, grazing, irrigated, continuously cropped, and forest. FLU estimates of irrigated area were aggregated by county, and were available for the years 2009, 2011, 2013, 2015, 2017, 2018, and 2019. Agreement between U-Net, FLU, and NASS was assessed with slope and $r^2$ from linear regression. 

\section{Results}

The confusion matrix in Table \ref{tab:confusion_matrix} shows the performance of the median model on the validation set.

\begin{table}[h]
\centering
\ra{1.3}
\begin{tabular}{rrrrrr}
& & & \multicolumn{2}{c}{$\textbf{Predicted}$} & \phantom{abc} \\
\cmidrule{3-6}
& & $\textbf{Irrigated}$ & $\textbf{Unirrigated}$ & $\textbf{Uncultivated}$ & \textbf{Recall} \\
 & \textbf{Irrigated} &1788615 &184405 &104779 & 0.86 \\
$\textbf{Actual}$ & $\textbf{Unirrigated}$  & 114903 &104929537 &1212801 & 0.99\\
 & \textbf{Uncultivated} & 194506 &5334847 &105872279 & 0.95 \\
 & \textbf{Precision} &0.85 & 0.95 & 0.97 &  \\
\bottomrule
\end{tabular}
\caption{Confusion matrix for U-Net predictions on validation set.}
\label{tab:confusion_matrix}
\end{table}

Over all years considered, U-Net had a f1-score for the irrigated class of 0.86 and 0.97 for both the uncultivated and the unirrigated class. There were differing numbers of irrigated labels in different years. The year 2003, for example, only had $\sim 3\times 10^4$ irrigated pixels and over $2.7\times 10^7$ non-irrigated pixels. The f1-score for 2003 was only 0.4, but for years with less imbalanced distributions (like 2013, with $\sim 3.5\times 10^5$ irrigated pixels)  the f1-score jumped to 0.87. Despite the fact that the f1-score was variable depending on the number of pixels labeled as irrigated per year, the model predictions looked similar throughout the time period evaluated.

A quantitative comparison of the performance of the different mapping techniques is shown in Table \ref{tab:value_added}. Metrics are only calculated for the irrigated class.
\begin{table}[h]
\centering
\ra{1.3}
\begin{tabular}{ccrrrr}
 & & $\textbf{OA}$ & $\textbf{P}$ & $\textbf{R}$ & \textbf{F1} \\
\cmidrule{3-6}
 & {IrrMapper RF} & 0.993 & 0.63 & 0.88 & 0.73 \\
\multirow{2}{0.10\columnwidth}{\textbf{Method}} & LANID RF  & 0.988 & 0.49 & 0.84 & 0.61 \\
 & {U-Net} & 0.997 & 0.87 & 0.88 & 0.88 \\
\bottomrule
\end{tabular}
\caption{Overall accuracy, precision, recall, and f1 score for the four different maps of irrigation in Montana. Precision, recall, and f1 are only reported for the irrigated class. Statistics were aggregated for the years 2008-2013 for LANID, IrrMapper, and the U-Net ensemble.}
\label{tab:value_added}
\end{table}

The U-Net ensemble outperforms existing maps of irrigation in Table \ref{tab:value_added}. There is a slight discrepancy between f1 scores in Tables \ref{tab:confusion_matrix} and \ref{tab:value_added} because labels from years 2003 and 2015 are omitted in Table \ref{tab:value_added} to keep the comparison between methods based on the intersection of the set of years where each product is available. The results in Table \ref{tab:value_added} aren't surprising, as U-Net was trained exclusively on data from Montana. However, it does show that U-Net is capable of classifying irrigated lands with higher precision and recall than other methods. All methods have an OA of greater than 98$\%$, reflecting the imbalanced nature of the validation data \footnote{Labeling each pixel in the validation set as unirrigated or uncultivated would still result in an OA of around 98\%}. For other states, especially states where irrigated agriculture comprises a larger percentage of land cover, overall accuracy may be a better estimate of model performance. However, the distribution of land cover classes by area in Montana is imbalanced and our validation set captures the distribution better than artifically balanced validation sets. The low precisions in Table \ref{tab:value_added} for LANID and IrrMapper are a result of both models labeling lots of dryland agriculture as irrigated.

\subsection{Validation with NASS and FLU}

\begin{figure}[h]
    \centering
    \includegraphics[width=\textwidth,height=\textheight,keepaspectratio]{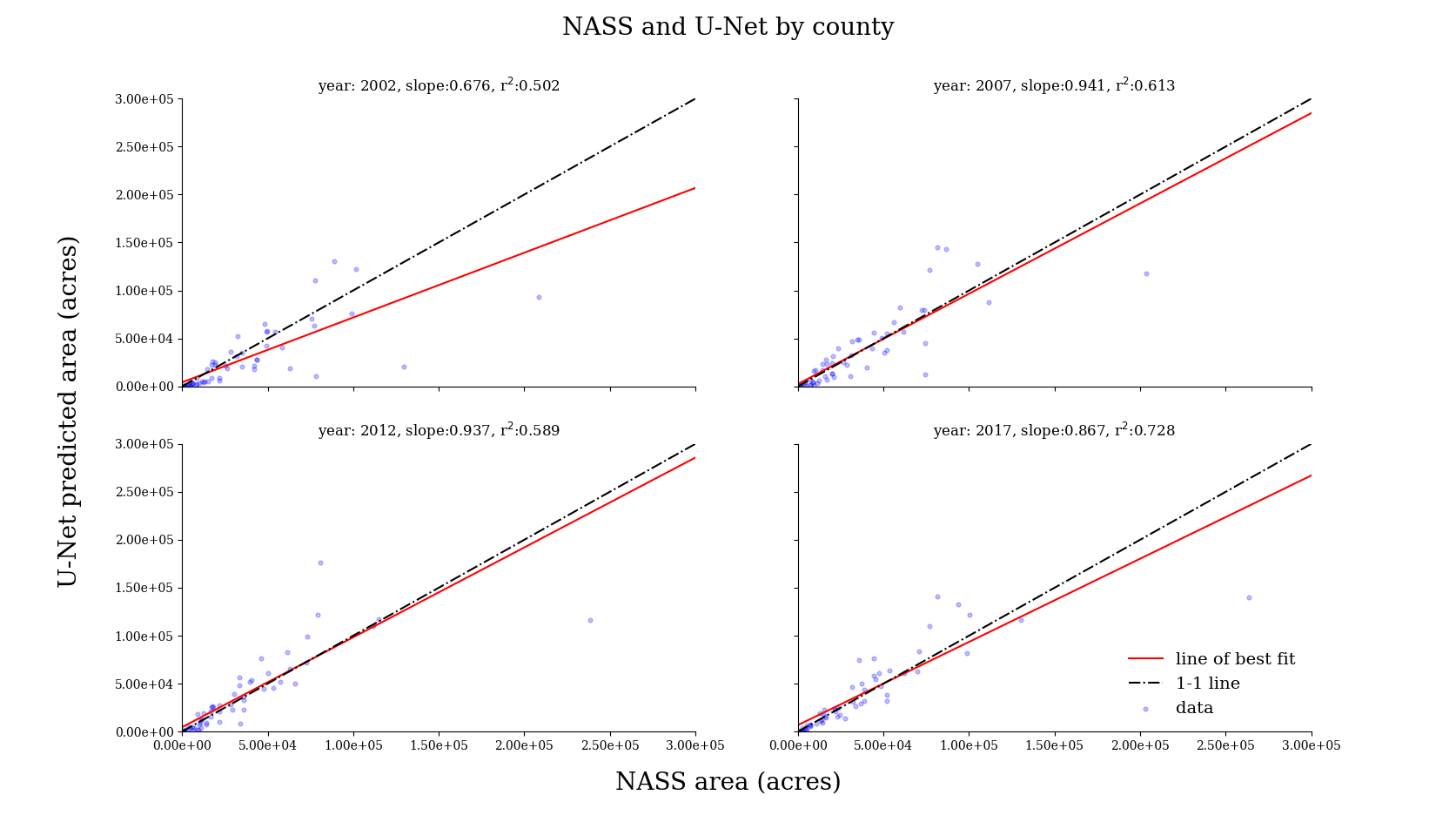}
    \caption{U-Net results compared to NASS statistics. The blue line on the left hand side indicates the line of best fit for the dataset, and the black line the 1-1 line.}
    \label{fig:unet_vs_nass_by_county}
\end{figure}

\begin{figure}[h]
    \centering
    \includegraphics[width=\textwidth,height=\textheight,keepaspectratio]{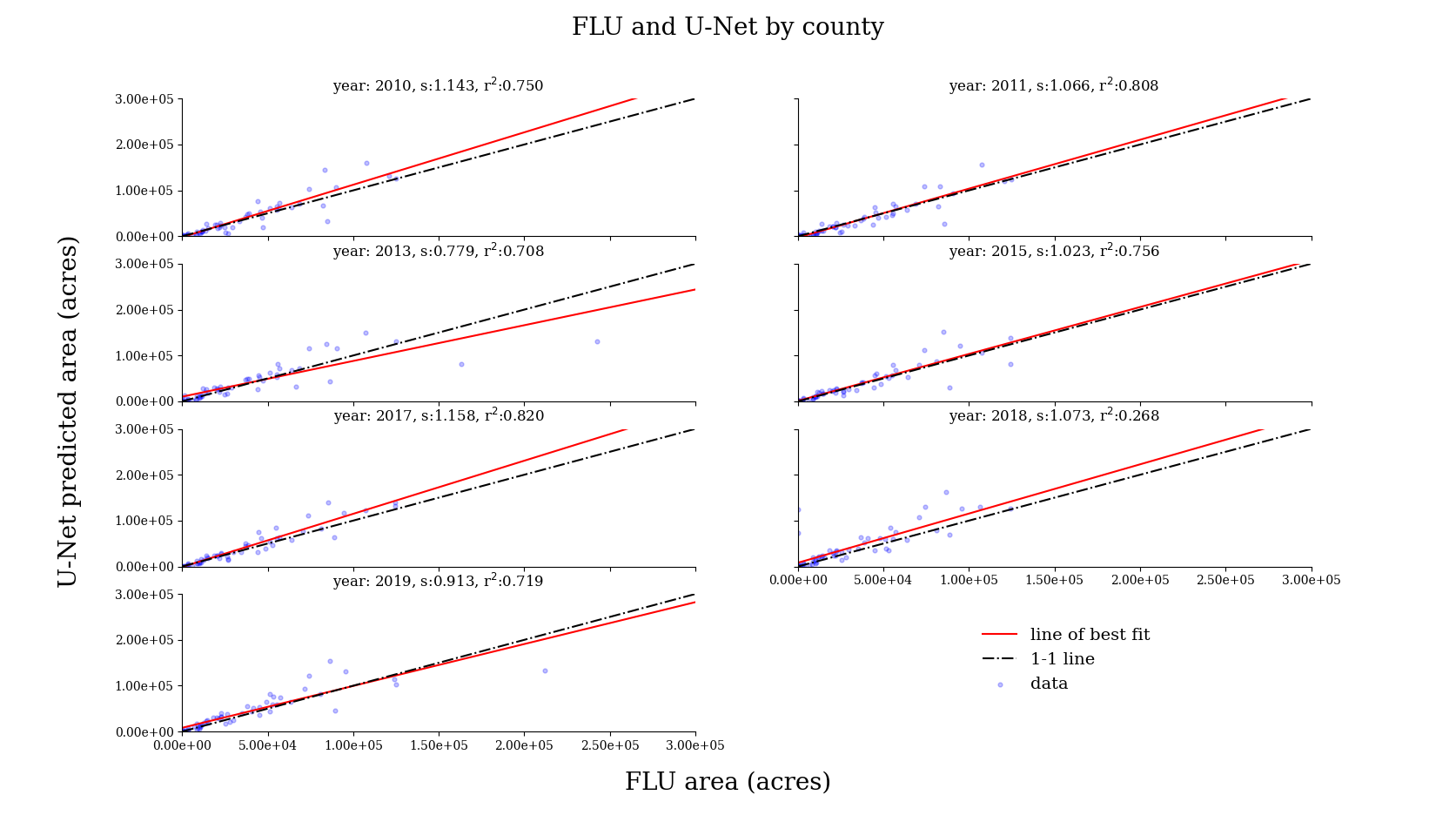}
    \caption{U-Net results vs FLU statistics.}
    \label{fig:unet_vs_flu_by_county}
\end{figure}

Fig. \ref{fig:unet_vs_nass_by_county} shows that U-Net slightly underestimates irrigated area by county when compared to NASS. Fig. \ref{fig:unet_vs_flu_by_county} indicates that U-Net and FLU data agree fairly well.

\begin{figure}[h]
    \centering
    \includegraphics[width=0.9\textwidth]{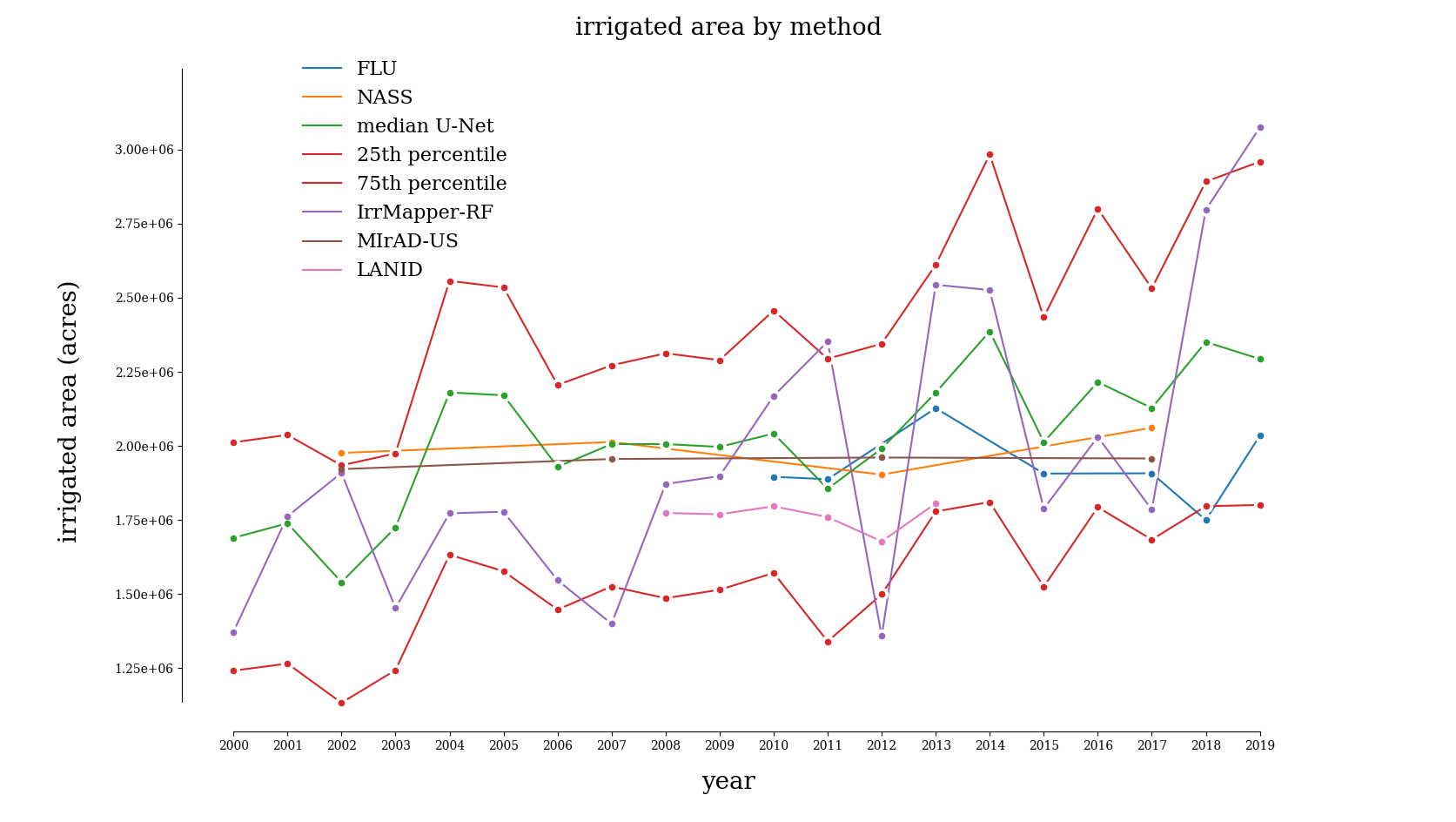}
    \caption{Irrigated area in Montana by method, with NASS and FLU data included. 25th and 75th percentiles for U-Net predictions are shown in red.}
    \label{fig:irrigated_area_by_method}
\end{figure}

U-Net estimates more irrigated area compared to FLU data on a county wide level for most of the years compared. Both comparisons show that variance in the comparison between census data (either FLU or NASS) increases with an increase in county irrigated area. The total area of irrigated agriculture predicted by the different methods is shown in Fig. \ref{fig:irrigated_area_by_method}. Overall, the median U-Net prediction is closer to both FLU and NASS than the two other methods examined (LANID and IrrMapper-RF). MIrAD-US agrees very well with NASS, as the method matches its predictions directly to NASS statistics.

\begin{figure}[ht]
    \centering
    \includegraphics[width=0.9\textwidth]{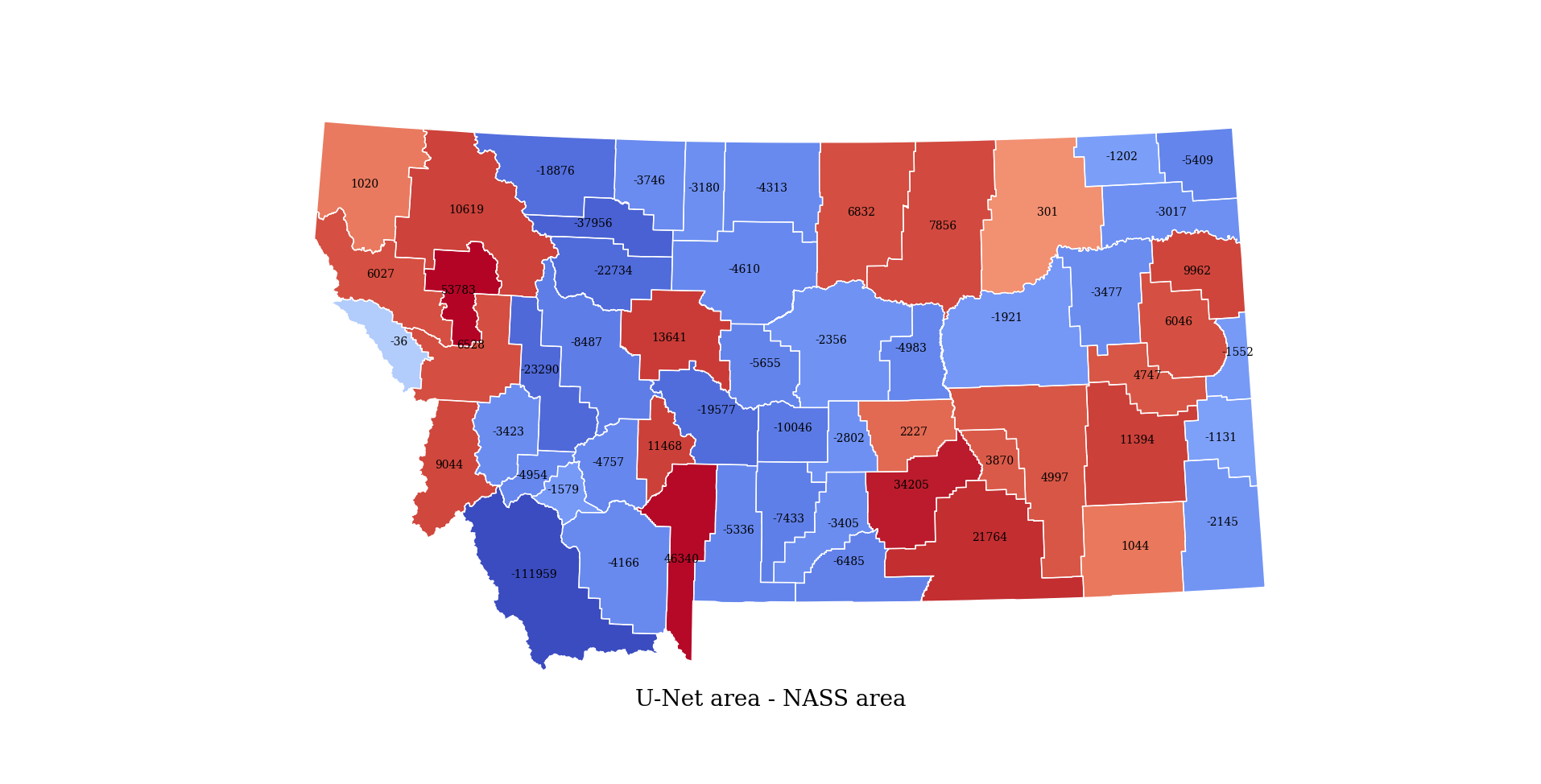}
    \caption{Mean difference in U-Net and NASS area. Mean taken over NASS census years.}
    \label{fig:mean_difference_unet_and_nass}
\end{figure}

There is a general increasing trend in irrigated area predictions in all methods of estimated irrigated area. IrrMapper-RF and U-Net agree in the trends in irrigated acreage that they identify. IrrMapper-RF predicted irrigated area follows trends in precipitation, most likely due to a) confusion in wet areas due to overall higher NDVI in wet years and b) the detection of temporary increases in irrigated land because of increased water availability. U-Net does not show the same correlation with precipitation as IrrMapper. In both comparisons to NASS and FLU, $r^2$ values vary widely, from 0.50 (2002) to 0.73 (2017). The large discrepancy between U-Net and NASS area for certain counties in 2002 is attributable to model error in counties on the Rocky Mountain Front. The model erroneously classified areas that it consistently predicts as irrigated in other years as unirrigated. We found that clouds were present in every image save one, meaning the model had little to no information to make classification. The clear outlier in all subplots in Fig. \ref{fig:unet_vs_nass_by_county} is Beaverhead county, in the bottom right of all plots. NASS reports irrigated area from 200,000 to 260,000 acres, while U-Net reports area from 90,000 to 140,000 acres. FLU data agrees very well with U-Net results for Beaverhead county, with the smallest difference in irrigated area prediction only 2,000 acres. U-Net consistently produces larger estimates of irrigation for a cluster of counties that NASS reports to have around 100,000 acres - Lake, Yellowstone, and Gallatin counties (Fig. \ref{fig:mean_difference_unet_and_nass}). FLU estimates of area for the same counties agree closely with U-Net estimates of irrigated area (15,000 acre difference for 2017). 

\section{Discussion}
\subsection{Qualitative Performance of Models}
Inspection of Landsat and NAIP imagery for NASS census years of Beaverhead county indicates that the large discrepancy in irrigated area between NASS and U-Net predictions is most likely due to U-Net omissions of flood irrigated lands as irrigated. U-Net tends to classify flood irrigation or irrigation at the wetlands-agriculture interface as wetlands. It's often hard to tell where wetlands begin and irrigation ends when inspecting high resolution satellite data, so it's not surprising that U-Net has difficulty distinguishing these two classes. In addition, in our training data and our study area, the existence of wetlands and irrigation in the same place is possible, and therefore both semantic and physical distinction between irrigated land and wetlands is blurred. This problem may be resolved by more extensive vetting of wetlands training data.

Each model has different biases on wetlands. LANID-2012 restricts its classification of irrigated land to land that is not wetlands, impervious surfaces, open water, and forests according to the 2012 CDL. Some flood irrigated land is classified as wetlands in the CDL, resulting in errors of omission. This is evident when inspecting NAIP imagery in the regions that the CDL classifies as wetlands. IrrMapper-RF, on the other hand, tends to label irrigated land that LANID and U-Net miss as irrigated. However, IrrMapper-RF also tends to label true wetlands as irrigated, for example clearly uncultivated land on islands in rivers. U-Net seems to bridge the gap between the two methods; it labels true wetlands as wetlands and some flood irrigated land as irrigated. However, U-Net doesn't detect as much flood irrigated land as IrrMapper-RF, resulting in lower estimates of irrigation in counties such as Beaverhead. Some wetlands training data incorporate pixels where flood irrigated land was labeled as wetlands. It's conceivable that the labels were correct at the time of their creation but the land use changed during subsequent years. In general, we expect IrrMapper-RF to overestimate irrigated land near wetlands-agriculture interfaces, U-Net to underestimate somewhat, and LANID to almost entirely ignore. This is evident in the Big Hole Valley near Wisdom, MT where LANID omits irrigated land. Classification of cultivated land at the wetlands-agriculture interface is difficult due to the possible semantic and physical overlap between irrigation and wetlands. Other land classes with less label uncertainty are accurately classified by U-Net compared to other methods. 

Dryland agriculture, is very well classified by U-Net compared to other methods, with few errors of commission. The errors mosty happen at the edges of irrigated fields that are labeled as dryland due to the difference in label and data resolution. LANID has many errors of commission in dryland areas, particularly in the eastern half of Montana. IrrMapper-RF also overpredicts irrigation in dryland agriculture but has fewer errors of commission in dryland areas when compared to LANID. The ease with which U-Net distinguishes irrigated and rainfed agriculture is one of the large advantages of using this product. 

U-Net systematically overestimates irrigation compared to NASS for Lake, Gallatin, and Yellowstone counties. These are all counties with large areas of irrigated land that reside next to bodies of water, and as such have large irrigation schemes. There are clearly fallowed fields that U-Net is confident are irrigated in some years. Fallow fields were not well represented in the training data compared to irrigated land. U-Net misclassifications in these areas could be solved by the creation of a larger fallow training set.

\subsection{Uncertainty in model predictions from prediction IQR}

The models in the ensemble did not always predict similar probabilities for each pixel. Histograms of the IQR for misclassified pixels are shown in Fig. \ref{fig:iqr_histograms}. The distributions of IQR for errors of commission on the irrigated class (whether on fallowed land, wetlands, dryland agriculture, or forest) are much different than the distributions for errors of commission on other classes. In particular, the pixels that are known to be non-irrigated yet predicted as irrigated tend to have distributions of IQR that peak in the 100-200 range. The 100-200 are quantized softmax probabilities that roughly correspond to a raw probability IQR of 0.4-0.8. The ability of the model ensemble to indicate some uncertainty makes directly fitting NASS or FLU data estimates of area by county possible by considering pixels irrigated that are above or below some IQR threshold. Land near the wetlands-agriculture interface often has a higher IQR than land that is clearly irrigated, indicating uncertainty in classification. Similarly, some fallow land classified as irrigated has a high IQR. The model ensemble also exhibits high IQR in mountainous regions of northwestern Montana that are erroneously classified as irrigated. 

\begin{figure}[h]
    \centering
    \includegraphics[width=0.9\textwidth]{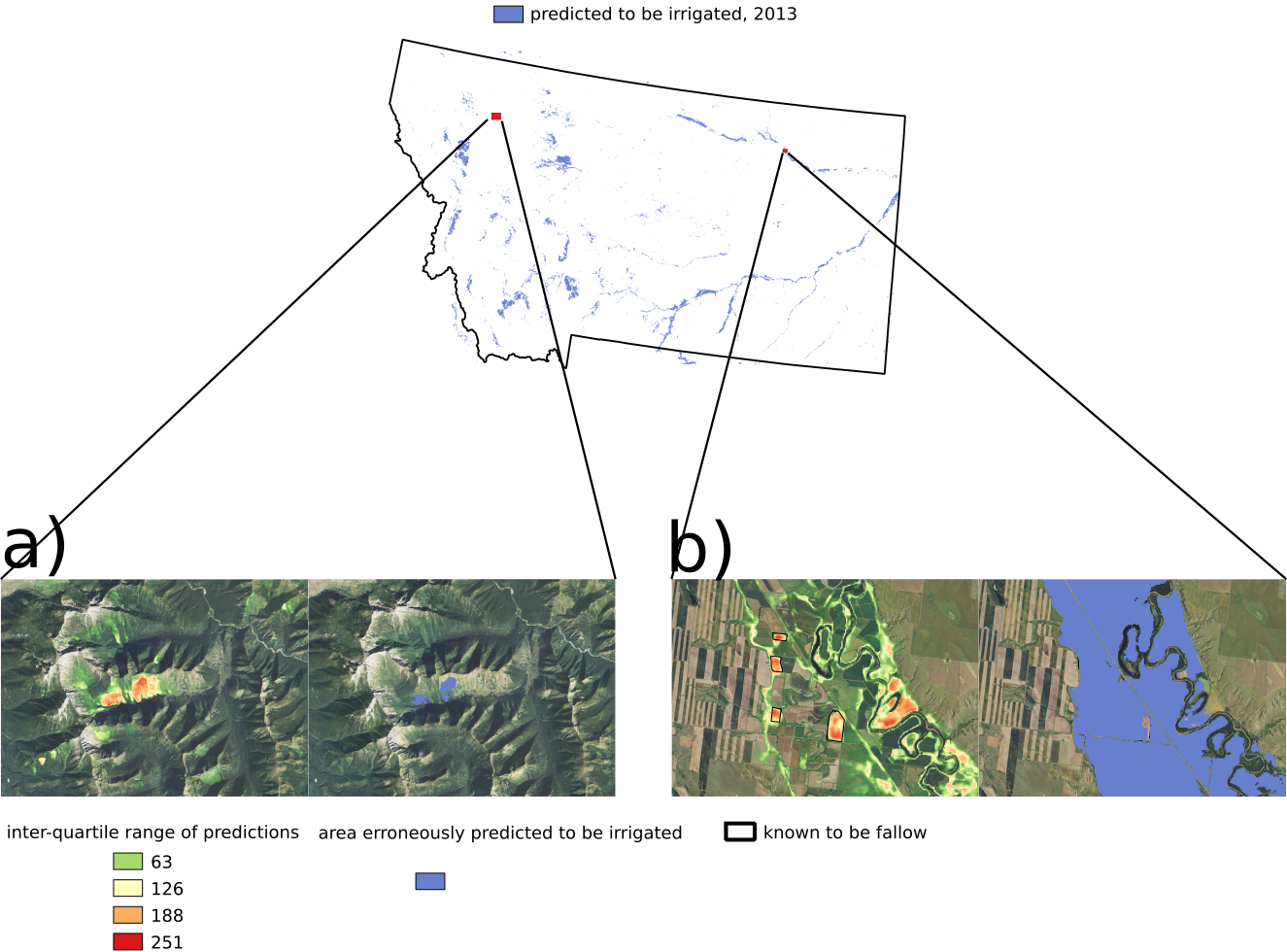}
    \caption{Example of large IQR where the model has incorrectly classified forest and fallowed lands as irrigated. IQR ranges from 0-255 instead of 0-1 because of prediction quantization. Best viewed in color.}
    \label{fig:uncertainty}
\end{figure}

The histograms of IQR for all pixels that the models predicted as a certain class are shown in Fig. \ref{fig:iqr_histograms_all_pixels_predicted}. Pixels predicted to be irrigated have a larger median IQR than pixels predicted as other classes. Even removing the first 30 bins (that contain most of the counts for the non-irrigated classes) results in median IQR of 5 and 8 for the unirrigated and uncultivated class, respectively. Median IQR for the irrigated class was 29. Examples of misclassifications that are highly uncertain are shown in Fig. \ref{fig:uncertainty}. Subfigure a) shows

U-Net predicts irrigation in some small regions in the mountains of northwest Montana. The IQR associated with the predictions clearly indicates that the ensemble is uncertain about the irrigated classification. Subfigure b) shows similar results for land that is known to be fallow - the fields outlined in black (which are known to be fallow but labeled by the ensemble as irrigated) clearly have a higher IQR. 

The discussion of the qualitative performance of U-Net results relied on maps of irrigation that assigned the median prediction of all 10 models to a given pixel. The 10 models perform differently because they were trained on different subset of the training data. In general, models that performed poorly on the validation set tended to have a less ``strict'' definition of irrigation and correctly predicted irrigation at the wetlands-agriculture interface more frequently than the median prediction. In contrast, models that performed well on the validation set tended to have a more restrictive definition of irrigation, and correctly identified more fallowed or dryland fields as irrigated in counties like Lake or Yellowstone. Part of this trend may be explained by the sparsity of labels in the validation set, as models that learn a very restrictive definition of irrigation may not truly reflect reality despite performing well on the validation set. All rasters used in the final analysis are available on GEE, and we recommend examining predictions qualitatively and examining model uncertainty when performing analysis.

\begin{figure}[h]
    \centering
    \includegraphics[width=0.9\textwidth]{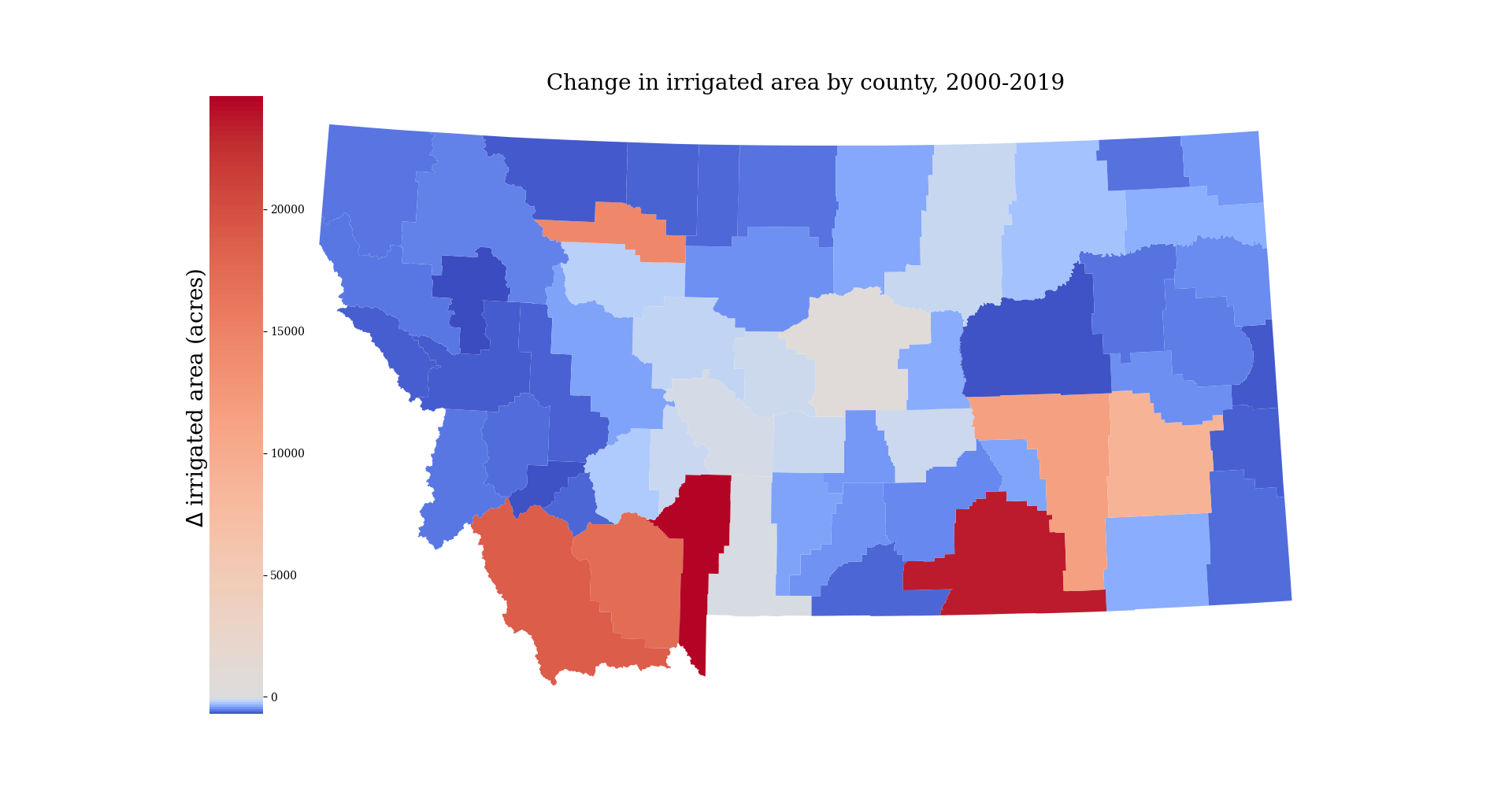}
    \caption{Change in irrigated area by county for the years 2000-2019. To smooth predictions, areas were averaged over 5 year non overlapping periods, and the year 2002 was excluded due to missing data issues.}
    \label{fig:change_in_area_by_county}
\end{figure}

The high accuracy and resolution of U-Net maps allow for a detailed accounting of irrigation in Montana. Overall, U-Net results indicate that irrigated area increased by 603,761 acres from 2000 to 2019. To reduce yearly noise, we averaged the results over 4 non-overlapping 5-year periods, finding that the averaged area over this time increased by 346,141 acres. The year 2002 was excluded from the analysis because of the issues due to missing data. Fig. \ref{fig:change_in_area_by_county} shows the change in irrigated land between the years 2000 and 2019. Many counties experienced a slight decrease in irrigation, while four strongly increased their irrigated land (Beaverhead, Gallatin, Big Horn, and Madison counties). 

Despite the fact that U-Net misclassifies flood irrigated land as wetlands in some areas in Beaverhead county, there is still a visible continuous increasing trend in irrigated agriculture.

\section{Conclusion}

We present a new methodology of mapping irrigation (U-Net) at high spatial resolution and over large regions in time and space, applied to the state of Montana. The method used is a deep learning model that ingests temporal mean images of satellite derived spectral data from multiple times throughout the year. It learns to mask clouds and missing data without explicit guidance. The model is also computationally efficient enough to produce predictions for 20 years over the state of Montana on a personal computer in less than a day. Our method agrees better statewide with both NASS and FLU data than existing maps of irrigation in Montana, and outperforms other methods on a hand-labeled validation set in overall accuracy, f1-score, and precision. The misclassifications of our method come from underpredicting irrigation at the wetlands-agriculture interface and overpredicting irrigation in fallow fields. Despite these drawbacks, our model clearly outperform existing methods in rainfed areas, making almost no errors of commission in dryland agriculture. The method we proposed is the first application of CNNs to mapping irrigation on a large spatial and temporal scale.

\printbibliography

\section*{Supplementary figures \& Appendix}

\begin{figure}[ht]
    \centering
    \includegraphics[width=0.9\textwidth]{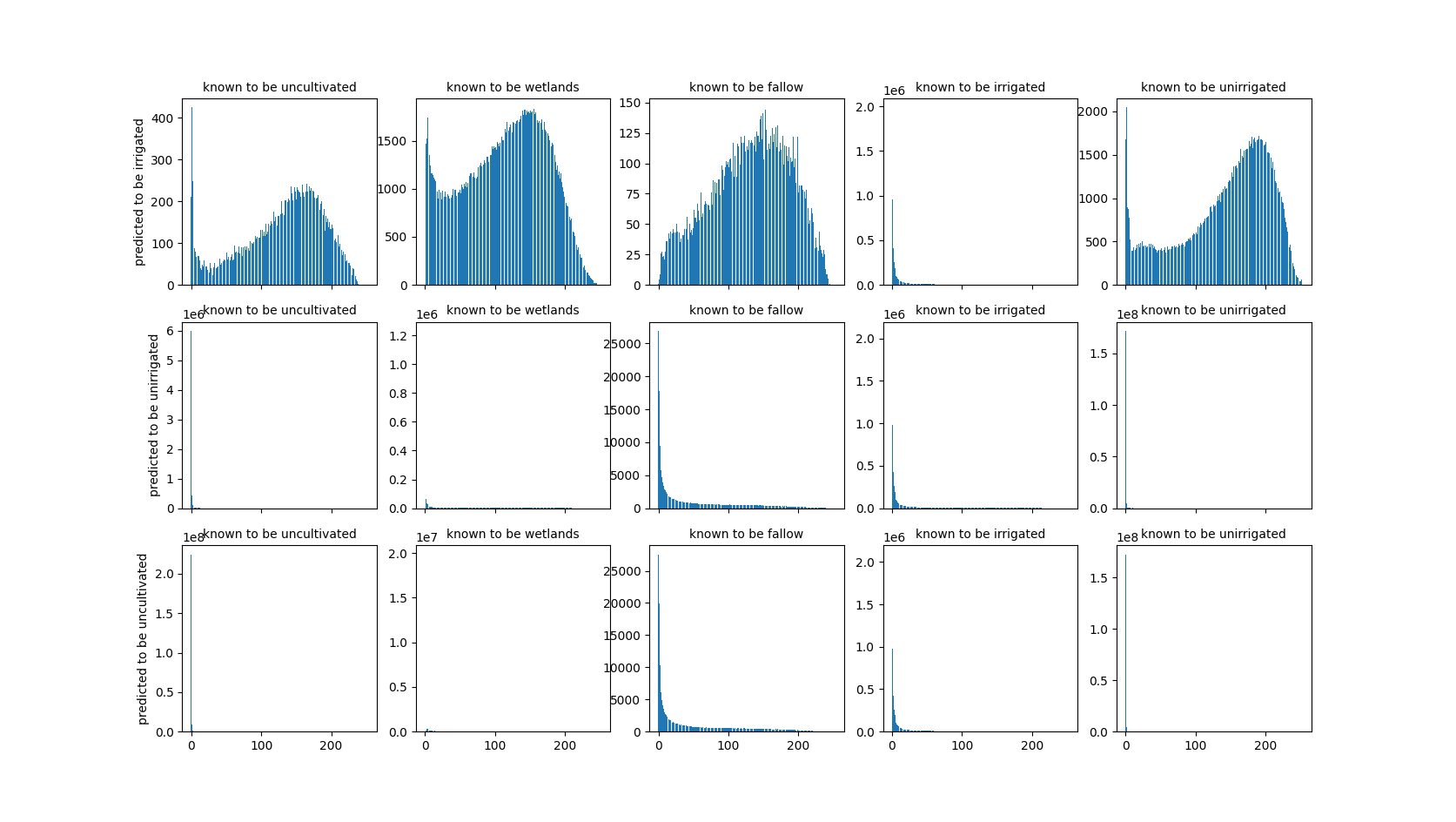}
    \caption{Histograms showing the IQR for correctly and incorrectly classified pixels.}
    \label{fig:iqr_histograms}
\end{figure}

\begin{figure}[ht]
    \centering
    \includegraphics[width=0.9\textwidth]{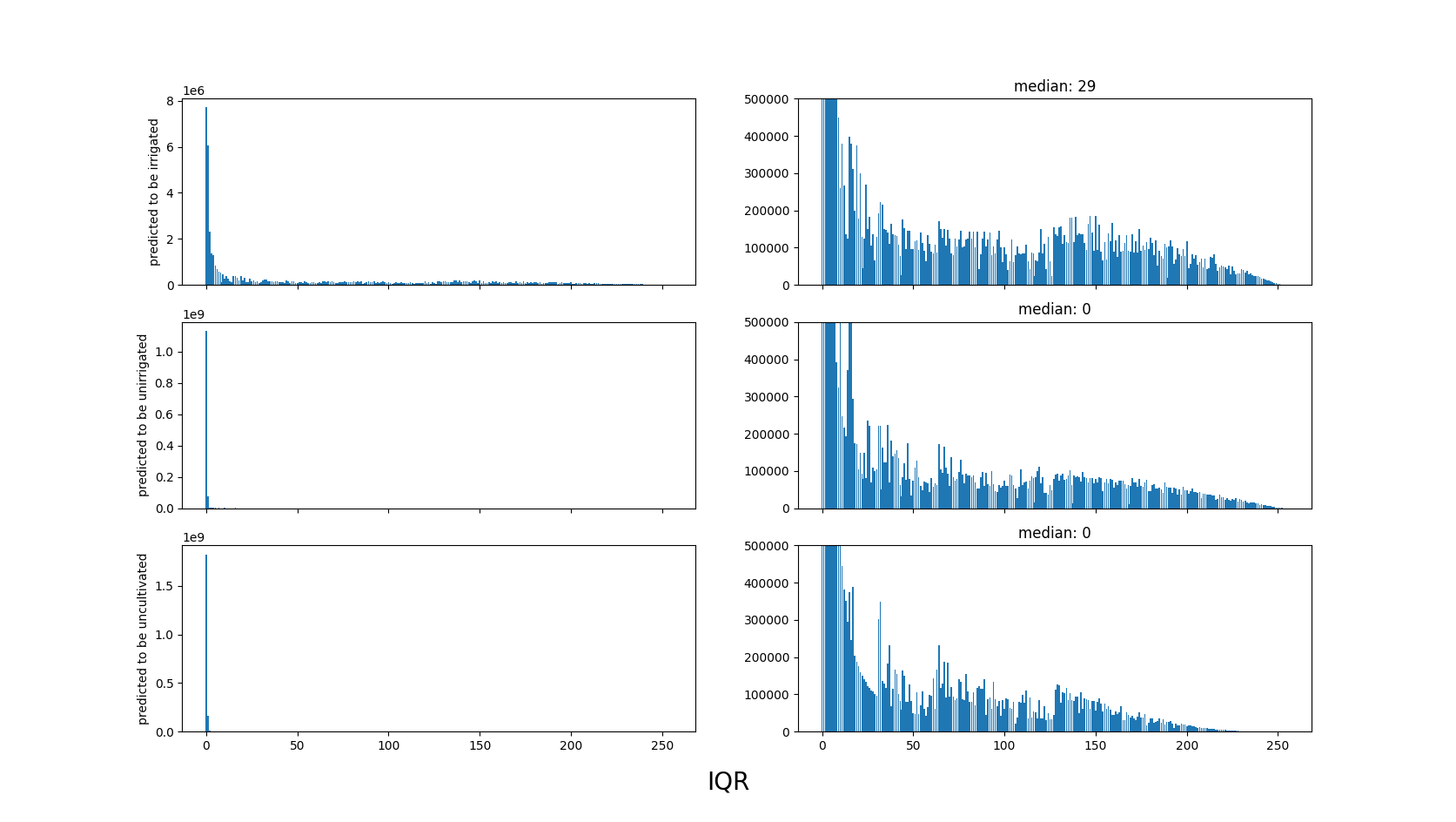}
    \caption{IQR histograms of all pixels predicted to be a given class. The right hand side of the figure shows a view where the y-axis has been restricted. Note: show cumulative distribution on a twinx()}
    \label{fig:iqr_histograms_all_pixels_predicted}
\end{figure}

Figures \ref{fig:iqr_histograms}, \ref{fig:iqr_histograms_all_pixels_predicted} show the histograms of the inter-quartile range (IQR) of model predictions for two different scenarios, on one year of labeled data. Fig. \ref{fig:iqr_histograms} shows the IQR of correctly and incorrectly classified pixels for each class. Overall, the IQR distributions for incorrect predictions of irrigation (row 1 in Fig. \ref{fig:iqr_histograms}) are much different than incorrect predictions of other class, peaking around 200.

The maximum number of pixels in each class is vastly different, shown in different y-axis scales. A zoomed-in version of these plots is shown on the right-hand side of Fig. \ref{fig:iqr_histograms_all_pixels_predicted}. For ease of visualization, we did not distinguish between wetlands, fallow, unirrigated, or uncultivated, instead aggregating all of these classes into one: non-irrigated.
The left hand side of the Fig. \ref{fig:iqr_histograms_all_pixels_predicted} shows the extreme skew in the data, with the vast majority of predictions having an IQR close to 0. On the right hand side, there is greater uncertainty in pixels predicted to be irrigated, regardless of whether or not they're predicted correctly. 

\subsection*{Appendix: Convolutional Neural Networks}

Convolutional neural networks (CNNs) are a class of statistical learning algorithms often applied to images. The core concept behind CNNs is the convolution of the input image with many convolutional ``filters''. A convolution of an image with a filter is analogous to densely sliding the filter over the image and recording the magnitude of the dot product of the filter (often a region of 3x3 pixels) with the small sub-region of the image. A discrete convolution is defined as in Eq. \ref{eq:convolution}.
\begin{equation}
    y[i, j] = \sum_{1}^{c} \sum_{-m}^{m} \sum_{-n}^{n} h_c[m, n] * x_c[i-m, j-n]
\label{eq:convolution}
\end{equation}
$y, x, h$ are the output image, the input image, and the convolutional filter, respectively. $m, n$ are integers indicating the width of the convolutional filter. For a filter of size 3x3, $m,n=1,1$, and for size 5x5 $m,n=2,2$. $c$ indicates a sum over channels. Each pixel in the output image is therefore a linear combination of multiple pixels in each channel of the input image depending on the kernel size. Eq. \ref{eq:convolution} represents a single convolutional filter and is densely applied across every pixel in the input image, with edge pixels either being discarded (``valid'' convolution), or incorporated through padding (often an image is padded with zeros to keep the size of the pre and post-convolution image the same). Often, many filters are applied to an image per layer, with common numbers in the range 32-1024. These filters build up a rich representation of the semantic content of an image, making classification of images into different classes simple. In order to build up a high-level description of the entire image, CNNs rely on downsampling steps to aggregate information on large spatial scales. 

Various options exist for downsampling images in a CNN, with a common one being 2-dimensional max-pooling. 2-D max pooling downsamples an image by creating a new image out of predetermined regions in the input image. For example, max pooling with size 2 and stride 2 computes the maximum of every non-overlapping 2x2 patch in the input image and uses the maximum to construct a new image that only contains the maximum of each 2x2 input patch, and is 1/4 the size. Max pooling can be applied with any size and any stride, but common choices are size 2 and stride 2. This downsampling steps reduce the size of the image, which increases computational efficiency and lets successive convolutional filters access larger regions of the input image. Part of the power of CNNs is their ability to learn non-linear transforms of the input image that make classification a simple task. This is encoded directly into the CNN model by the application of nonlinearities, of which the rectified linear unit (ReLU) is a common choice. Non-linearities were designed with the observation that certain signals are amplified and others dampened within synapses in a brain. Non-linearities are typically applied after a filter bank. After many successive applications of convolutional filters, non-linearities, and downsampling steps, the input image has been transformed in such a way as to make classification easy. To this end, a classification layer (which simply maps the output of the last convolutional block to \textit{n} values, one for each class) is tacked on with a special activation function called softmax (Eq. \ref{eq:softmax}) The softmax function converts a sequence of numbers into values in the range 0 - 1 that are often interpreted as probabilities.
\begin{equation}
    \texttt{SM}(a_i) = \frac{\texttt{exp}(a_i)}{\sum_{j=1}^{n} \texttt{exp}(a_j)}
\label{eq:softmax}
\end{equation}
In Eq. \ref{eq:softmax}, $a_i$ is the softmax of element $i$. The maximum of the softmax output is considered as the class the CNN predicts a given image to be. At the beginning of training, the softmax output is often random, as the CNN hasn't been tuned for classification yet via a loss function.
 
A common loss function for classification is the categorical cross-entropy, which relies on the multinomial distribution to quantify the difference between the predictions of a CNN and ground-truth labels. The parameters of the CNN are updated via backpropagation, which propagates the error signal from the loss function backwards through every parameter of the network. Often, the algorithm used to determine how much of the error signal is backpropagated through a CNN is some variant of stochastic gradient descent (SGD). SGD tries to find a set of network parameters that minimizes the loss function by tweaking the parameters of the network proportional to the gradient of the loss function with w.r.t the parameters. CNNs are susceptible to overfitting due to their large numbers of parameters (often in the millions or even billion). To partially remedy this issue, various methods of regularization have been adopted to penalize pathological behavior of networks. Penalizing the $l_2$ norm of the parameters in a neural network is a common technique often called weight decay, motivated by the observation that large parameter magnitudes in linear regression are often a sign of overfitting.

\end{document}